\documentclass{article}

    \PassOptionsToPackage{numbers, compress}{natbib}

\usepackage{neurips_data_2023}
\usepackage{subcaption}

\usepackage[utf8]{inputenc} 
\usepackage[T1]{fontenc}    
\usepackage{hyperref}       
\usepackage{url}            
\usepackage{booktabs}       
\usepackage{amsfonts}       
\usepackage{nicefrac}       
\usepackage{microtype}      
\usepackage{xcolor}         
\usepackage[capitalize, noabbrev]{cleveref}
\usepackage{multirow}
\usepackage{graphicx}
\usepackage{subcaption}

\newcommand{\ie}{\textit{i.e.}}

\title{Sarcasm in Sight and Sound: Benchmarking and Expansion to Improve Multimodal Sarcasm Detection}

\author{
Swapnil Bhosale\\
University of Surrey, UK
\And
Abhra Chaudhuri\\
University of Surrey, UK
\And
Alex Lee Robert Williams\\
University of Surrey, UK
\And
Divyank Tiwari\\
IIT Bombay, India
\And
Anjan Dutta\\
University of Surrey, UK
\And
Xiatian Zhu\\
University of Surrey, UK
\And
Pushpak Bhattacharyya\\
IIT Bombay, India
\And
Diptesh Kanojia\\
University of Surrey
}

\begin{document}
\nolinenumbers
\maketitle

\begin{abstract}
The introduction of the MUStARD dataset, and its emotion recognition extension MUStARD++, have identified sarcasm to be a multi-modal phenomenon -- expressed not only in natural language text, but also through manners of speech (like tonality and intonation) and visual cues (facial expression).
%
With this work, we aim to perform a rigorous benchmarking of the MUStARD++ dataset by considering state-of-the-art language, speech, and visual encoders, for fully utilizing the totality of the multi-modal richness that it has to offer, achieving a 2\% improvement
in macro-F1 over the existing benchmark.
%
%
Additionally, to cure the imbalance in the `sarcasm type' category in MUStARD++, we propose an extension, which we call \emph{MUStARD++ Balanced}, benchmarking the same with instances from the extension split across both train and test sets, achieving a further 2.4\% macro-F1 boost.
The new clips were taken from a novel source -- the TV show, House MD, which adds to the diversity of the dataset, and were manually annotated by multiple annotators with substantial inter-annotator agreement in terms of Cohen's kappa and Krippendorf's alpha.
Our code, extended data, and SOTA benchmark models are made public.
\footnote{Contact: \textit{s.bhosale@surrey.ac.uk, d.kanojia@surrey.ac.uk}}
\end{abstract}

\section{Introduction}
Sarcasm is a manner of speech that conveys a sense of contempt or mockery through the use of irony. For instance, the remark: \emph{"The tropical summers are so pleasant!"}, although it refers to tropical summers as being pleasant, it is, in fact, expressing contempt for the extreme temperatures experienced in the tropics during summers. However, not all sarcastic remarks are expressed solely through natural language. For instance, the statement: \emph{"The cake is so delicious!"}, may or may not be sarcastic. If it were sarcastic, there would be no way to say so unless one looked at the speaker's facial expression, and paid attention to the tonality of the remark. Such instances pose a challenge for the automatic detection of sarcasm using only text as input. Further, it is revealed that sarcasm detection is an inherently multi-modal problem involving natural language, speech, and vision~\cite{Castro2019Mustard, ray2022Mustardpp}.
Recently, many other natural language processing (NLP) tasks under the \textit{sentiment umbrella} have been investigated through the lens of multimodality~\cite{soleymani2017survey,kiela2020hateful,suryawanshi2020multimodal,guo2023audio}. 
Sarcasm detection is usually performed as a supervised learning-based classification task, and hence, requires a significant number of hard-to-obtain labelled multimodal data. These challenges motivate our investigation into the study of multi-modal sarcasm detection, and further the efforts with the help of additional data and rigorous benchmarking.

With this motivation, MUStARD~\cite{Castro2019Mustard}, a multi-modal sarcasm detection dataset, was introduced; and it provides information from all three associated modalities, \ie, text, speech, and video. Additionally, Ray \textit{et al.}~\cite{ray2022Mustardpp} recognized underlying emotions to be a crucial determinant of the sarcastic nature of remarks, and introduced MUStARD++, an extension of MUStARD with emotion annotations, and an expansion on the number of instances.
The associated benchmarking performed in the above works leverage the multi-modal nature of their datasets by encoding the individual modalities through independent text, speech, and vision modules, and combining their respective representations via a fusion layer~\cite{liu2019use}.
We identify two limitations in the existing multi-modal sarcasm detection benchmarking literature. First, all benchmarks on the task have so far only relied on classical encoding approaches of the individual modalities. For instance, MUStARD++, which is one of the most recent benchmarks in the literature, used Mel Frequency Cepstral Coefficients (MFCC), Mel spectrogram (using the Librosa library~\cite{McFee2015Librosa}), prosodic features from OpenSMILE~\cite{opensmile} for speech, and ResNet-152~\cite{He2016ResNet} for videos, which cannot fully leverage the complementarity in the multi-modal information available in the datasets.
Second, we also observe that the `sarcasm type' category in MUStARD++ exhibits an imbalance. Through this work, we aim to bridge these gaps by rigorous benchmarking based on SOTA vision, speech, and language encoders. We also aim to expand the MUStARD++ dataset by adding new instances that alleviate this imbalance to a certain degree.

Recent literature on self-supervised learning has shown that large-scale self-supervised pre-training can often generalize much better than even supervised pre-training.
The introduction of CLIP~\cite{radford2021clip} enabled this for multi-modal vision-language tasks as well.
Through this work, we propose to perform a benchmarking for the task of multi-modal sarcasm detection that is more in tune with the current state-of-the-art (SOTA) in multi-modal learning. We leverage ViFi-CLIP~\cite{rasheed2023fine}, a SOTA CLIP-based model for encoding videos in a common video-text representation space. We use Wav2vec 2.0, a transformer-based self-supervised speech representation learning model trained using a masked learning objective for audio. With our proposed benchmarking, we were able to achieve $>2\%$ gain in macro-F1 over the existing benchmark.

To cure the `sarcasm type' category imbalance in MUStARD++, we propose an extension, we call, \emph{MUStARD++ Balanced}. The publicly available clips in our extension are taken from the TV show, House MD, which, apart from curing the imbalance, also adds diversity to the dataset by virtue of being a novel source that had not been considered in any form either in MUStARD or MUStARD++. Our dataset extension has been annotated by three independent annotators who show a substantial inter-annotator agreement in terms of Cohen's kappa and Krippendorf's alpha. We split the clips in our extension across both train and test sets, and perform further benchmarking using our proposed setup. Our experiments revealed a further $2.4\%$ boost to macro-F1 from using our extension, and show that this balance, as well as, the expansion in terms of the number of data points, could be effectively leveraged by our benchmark models to achieve notable improvements in sarcasm detection performance. We perform additional experiments that quantify the distribution shift introduced by our extension, demonstrating the added diversity.

To summarize, we make the following \textbf{contributions}: \textbf{(1)} rigorous benchmarking for the task of multi-modal sarcasm detection on the MUStARD++ dataset using SOTA vision, speech, and language models; \textbf{(2)} an extension to the MUStARD++ dataset, that we call \emph{MUStARD++ Balanced}, to cure the `sarcasm type' imbalance in the former, and to add data diversity; and \textbf{(3)} extensive experiments that show the significant gains in macro-F1 obtained by using our proposed benchmarking strategy, as well as from using our novel, balanced extension.

\section{Related Work}

We discuss the literature related to sarcasm detection and multimodal learning in the subsections below.

\subsection{Sarcasm Detection}
Transformers~\cite{NIPS2017_3f5ee243} architecture-based approaches have increased in prevalence within NLP and also within sarcasm detection literature. This is most notably due to their ability to pick up semantic and syntactic relationships within text. Various rule-based and machine learning-based approaches to sarcasm detection have been discussed in~\cite{10.1145/3124420}, and they also present a linguistic perspective to sarcasm detection.~\cite{7549041} used a pattern-based approach for the task, while emphasizing the role of four sets of features obtained based on different sarcasm types. This pattern-based study achieved $83.1$\% accuracy and $91.1$\% precision on the task of sarcasm detection. Since the advent of Transformers, usage of machine learning approaches has seen a steep decline. Some studies include~\cite{reyes-rosso-2011-mining} and~\cite{barbieri-etal-2014-modelling}, which used a Naive Bayes and Decision Tree
model, respectively, in order to identify sarcasm
where both achieve the best F1 scores over 70 on
their chosen datasets. 

On the dataset released with the SemEval 2018 Shared Task 3~\cite{van-hee-etal-2018-semeval},~\cite{Potamias_2020} offered an RCNN-RoBERTa methodology, where a RoBERTa transformer was used with BiLSTM to enhance task performance. Further, they also report that the RCVV-RoBERTa approach achieved an F1-score of 90.0 on the Riloff dataset~\cite{riloff-etal-2013-sarcasm}. Several recent methods for sarcasm detection using text are discussed by~\cite{shangipour-ataei-etal-2020-applying}. A BERT~\cite{devlin2019bert} model without concatenated layers, BERT encodings with a Logistic Regression model, and other language models like IAN~\cite{ijcai2017p568} that are trained and assessed on a Twitter-based sarcastic dataset are among these more recent efforts. With an F1-score of 73.4 in those evaluations, the BERT language model, without any additional layers, performs the best. Some existing literature investigates methods for performing sarcasm
detection in Arabic~\cite{abu-farha-magdy-2021-benchmarking}, where an extensive set of experiments are performed on different transformer architectures, that include mBERT, XLM-RoBERTa~\cite{conneau-etal-2020-unsupervised} and language-specific models like MARBERT~\cite{abdulmageed2021arbert}. In this low-resource setting for Arabic, the most effective model in this study achieves an F1-score of 58.4, which shows the need to investigate sarcasm in a multilingual low-resource setting. A weighted average Ensemble of a CNN, LSTM, and Gated Recurrent Unit (GRU) based architectures is trained with GloVe~\cite{pennington-etal-2014-glove} word embeddings to identify sarcasm, as demonstrated in~\cite{goel2022sarcasm}. This Ensemble outperformed comparative studies by up to 8\% on SARC~\cite{khodak2018large}, a Reddit comments dataset.

\subsection{Multi-modal Learning}

Existing literature on multimodal sentiment classification refers to the MOUD~\cite{perez2013utterance} and MOSI~\cite{zadeh2016mosi} datasets, while the IEMOCAP dataset~\cite{busso2008iemocap} for the task of multimodal emotion recognition. Poria \textit{et. al.}~\cite{poria2017context} propose the use of a bidirectional contextual long short-term memory (bc-LSTM) architecture for both tasks and show improvements over baseline on all three datasets. However, Majumder \textit{et. al.}~\cite{majumder2018multimodal} later propose context modelling with a hierarchical fusion of multimodal features and achieve improved performance in a monologue setting. In the conversation setting, Hazarika \textit{et. al.}~\cite{hazarika-etal-2018-conversational} propose using a Conversational Memory Network (CMN) to leverage contextual information from the conversation history and achieve improved performance. Novel multimodal neural architectures~\citep{wang2019words,pham2019found} and multimodal fusion approaches~\citep{liang2018multimodal,tsai2018learning} have propelled the deployment of computational models in this domain, while more efficient multimodal fusion approaches have also been discussed in~\citep{sahay-etal-2020-low,tsai-etal-2019-multimodal,liu-etal-2018-efficient-low}.

For multimodal sarcasm detection, a recent survey discusses the datasets and approaches in detail~\cite{9753058}. The MUStARD dataset~\cite{mustard} provides clips compiled from popular TV shows, including Friends, The Golden Girls, The Big Bang Theory, and Sarcasmaholics Anonymous, annotated with sarcasm labels. Ray \textit{et. al.}~\cite{ray-etal-2022-multimodal} extend upon this dataset by adding emotion labels and additional clips while also benchmarking for the multimodal sarcasm detection task. They call this extended dataset \textit{MUStARD++} and utilise feature fusion~\cite{liu2019use} and a feed-forward network to predict the sarcasm label. The authors show an F1-score of $70.2$\% points using audio, text and video modalities.

\section{Dataset details}

The MUStARD~\cite{Castro2019Mustard} dataset contains a total of $6,365$ videos, collected from TV shows like Friends, The Golden Girls, The Big Bang Theory, and Sarcasmaholics, of which $345$ are sarcastic, and $6,020$ are non-sarcastic. Although some works in the literature have performed benchmarking on the MUStARD dataset~\cite{liang2021multibench, Castro2019Mustard, ray2022Mustardpp}, they do not consider the latest developments in multi-modal vision-language fusion~\cite{radford2021clip, rasheed2023fine} and speech processing~\cite{chen2023exploring}. Chauhan \textit{et al.}~\cite{Chauhan2020SarcasmEmotion} annotated MUStARD with 9 emotion labels, which was refined and extended by Ray \textit{et al.}~\cite{ray2022Mustardpp} in MUStARD++ by adding valence and arousal information, which are important indicators for recognizing the emotion behind a remark, as well as adding further data points.
\cref{tab:dataset_stats} summarizes the statistics of the above datasets, and compares it with our extension. MUStARD++ extended MUStARD (345 sarcastic video samples) with $256$ sarcastic video samples ($216$ clips from \textit{The Big Bang Theory} and $40$ from \textit{Silicon Valley}), and they label $601$ non-sarcastic instances out of $6,020$ in MUStARD to balance the annotation. Along with refining the emotion labels in~\cite{Chauhan2020SarcasmEmotion}, they annotated all their data points with information about \emph{valence}, the positivity/negativity of emotion, and \textit{arousal}, a measure of the associated emotional intensity~\cite{picard1997affective}.

\begin{table}[!ht]
    \centering
    \begin{tabular}{cccc}
         \toprule
         & \textbf{Sarcastic} & \textbf{Non-Sarcastic} & \textbf{Total} \\
         \midrule
         MUStARD & $345$ & $6020$ & $6365$ \\
         MUStARD++ & $601$ & $601$ & $1202$ \\
         \midrule
         \textbf{Post our extension [MUStARD++ Balanced]} & $691$ & $674$ & $1365$\\
         \bottomrule
    \end{tabular}
    \caption{Comparison of statistics of existing sarcasm detection dataset and our proposed extension.}
    \label{tab:dataset_stats}
\end{table}



\subsection{Dataset Extension}
\label{subsec:dataext}
MUStARD++ categorised their data points by adding the `sarcasm type' information. Specifically, they included the following sarcasm-type annotations to their dataset:
\begin{itemize}
    \item \textbf{Propositional (PRO):} Remarks that need additional context to be classified as being sarcastic. For example: "\textit{That's very kind of you.}"
    \item \textbf{Illocutionary (ILL): } The type of sarcasm where the irony is expressed through non-textual cues, like voice or gesture. For example, while tasting a poorly cooked food, one may remark, "How delicious!", while expressing their contempt through facial expressions.
    \item \textbf{Embedded (EMB): } When the sarcastic incongruity is clearly embedded in the text. For example, "\emph{What a pleasant orchestra of honking cars!}"
    \item \textbf{Like-Prefixed (LIKE):} Prefixes the statement with a "like" to stress the irony. For example, "\textit{Like it means the end of the world!}".
\end{itemize}

A lack of balance in a sarcasm detection dataset among the above types can make a model develop a bias in favour or against certain types of sarcasm, which would make the downstream system unfair. Thus, ensuring that the training dataset contains fair proportions of all four types of sarcasm is the first step towards building a trustworthy sarcasm detection system.
\cref{tab:proposed_stats} shows our attempt to balance the various sarcasm types in MUStARD++. It can be seen that MUStARD++ has a clear under representation of samples in the EMB and the LIKE types, when compared with the PRO and ILL types. As can be seen from the $\%$ Change column, we curate our extension such that the above imbalance can be somewhat mitigated.

\paragraph{Data Validation \& Analysis} Per our annotators, sarcasm annotation and its type are challenging to identify and cognitively cumbersome; thus, our annotation task spanned longer than anticipated. Multimodal data annotation is further challenging as it requires annotators to watch lengthy videos when compared to textual data annotation. Hence, our data annotation instances are also limited in number. However, with three annotators, our annotation shows a substantial agreement ($\mathbf{0.743}$) using Fleiss' Kappa ($\displaystyle K$), and ($\mathbf{0.798}$) Krippendorf's alpha ($\alpha$). All three annotators were graduate students who were paid an agreed-upon fixed compensation for the annotation task, and consent was obtained to use the collected data for research and public release. We extend MUStARD++ by a total of $164$ instances, out of which $91$ are sarcastic. For final instance labels, we choose the majority label among the three for sarcastic/non-sarcastic. There was no difference in the \textit{sarcasm-type} annotation among the majority of annotators for the final label. 

\paragraph{License \& Potential Negative Impact} We license our data extension\footnote{Data and Code: \url{https://bit.ly/3MVnZOj}}, raw annotations, and code under the CC-BY-NC-4.0\footnote{\url{https://creativecommons.org/licenses/by-nc/4.0/}} license, the \textit{same as MUStARD++}. 

Multimodal datasets are often collected using content produced for television, live shows, \textit{viz.,} stand-up comedy, podcasts, \textit{etc.} However, such content can often contain material which is offensive in nature, for the purpose of creating humour. Similarly, sarcasm is also used to generate humour emanating offence at times. We acknowledge that our data and instances from MUStARD++ possibly contain such material. We release this data for ease of reproducibility and for propagating further research in the area. For our work, we obtain data from publicly available sources material on Youtube and release it under the aforementioned license.  

\begin{table}[!ht]
    \centering
    \resizebox{\columnwidth}{!}{
    \begin{tabular}{ccccccc}
         \toprule
         Sarcasm Type & MUStARD++ & \textbf{Our Extension} & Combined & MUStARD++ \% & Combined \% & \% Change \\
         \midrule
         PRO & 333 & 37 & 370 & 55.408 & 53.545 & -1.862 \\
         ILL & 178 & 15 & 193 & 29.617 & 27.931 & -1.686 \\
         EMB & 87 & 35 & 122 & 14.476 & 17.656 & 3.18 \\
         LIKE & 3 & 3 & 6 & 0.499 & 0.868 & 0.369 \\ \hline
         TOTAL & 601 & \textbf{90} & \textbf{691} & 100 & 100 & - \\
         \bottomrule
    \end{tabular}
    }
    \caption{Statistics of our MUStARD++ Balanced extension, and its comparison with MUStARD++.}
    \label{tab:proposed_stats}
\end{table}
\section{Methodology}
We first describe the individual backbones proposed for experimentation with each modality, \ie ViFi-CLIP for video and text encoders, wav2vec 2.0 for audio, and BART for text, respectively. Particularly, in the case of the audio backbone, we present our results when extracting features pre- and post-fine-tuning the entire backbone on an auxiliary speech emotion recognition (SER) task.
\subsection{Video \& Text Backbone: ViFi-CLIP}
Rasheed~\textit{et al.}~\cite{rasheed2023fine} propose a new approach for video learning that uses fine-tuned CLIP models. CLIP is a large-scale pre-trained model which learns to map images and text to a shared embedding space- allowing it to learn visual and semantic features from images. ViFi-CLIP shows that fine-tuning CLIP models on a small dataset of videos can improve the performance of video learning models on a variety of tasks; particularly because fine-tuning allows the CLIP models to learn temporal information which is not the case with the original CLIP model. The authors evaluated their approach on a variety of video learning benchmarks, and it outperformed the existing state-of-the-art.

Given a video sample $V_i \in R^{T \times H \times W \times C}$ with $T$ frames, and corresponding text label $Y$, ViFi-CLIP's video encoder utilizes a CLIP image encoder to encode the T frames independently as a batch of images and produce frame level embeddings $x_i \in R^{T \times D}$. In order to incorporate temporal learning the frame-level embeddings are aggregated using an average-pooling operation to obtain a video-level
representation $v_i \in R^D$. During training, the authors maximize the cosine similarity between video embeddings from the video encoder and text embeddings obtained from CLIP's text encoder, for text corresponding to each video. We utilize ViFi-CLIP's video encoder (V1), and the text encoder (T1) together in combination and also separately for experimentation and reporting results. 

\subsection{Audio Backbone: Wav2vec2}
Self-supervised models exploit large scale speech corpus without explicit labels. Wav2vec2 is one such self-supervised model based on transformers; that adopts a masked learning objective to predict missing frames from the remaining context. Intermediate features from Wav2vec2 have been widely explored in the literature to extract contextualized features for related downstream tasks or even fine-tune the entire model on auxiliary tasks.  It consists of three sub-modules, feature encoder, transformer module, and quantization module. Feature encoder is a multi-layer CNN that processes the input
signal into low-level features. Based on this representation,
the transformer module is further applied to produce contextualized representation. The quantization module discretizes
the low-level features into a trainable codebook. To train the
model, part of the low-level features are masked from the
transformer module, and the objective is to identify the quantized version of the masked features based on its context.

\subsubsection{Fine-tuning Wav2vec2 on the SER task}
Speech emotion recognition and sarcasm recognition are both tasks related to understanding and interpreting the emotions and intentions behind the human speech. While they are not directly correlated, there can be some overlapping aspects in terms of feature extraction and contextual understanding. To infer the underlying emotion, SER involves analyzing various acoustic and linguistic cues, such as pitch, intensity, speech rate, prosody, and word choice. Similarly, sarcasm detection requires understanding the context, the speaker's intention, and subtle linguistic cues, such as intonation, emphasis, and contradictory statements, to recognize sarcastic remarks.

We hypothesize that fine-tuning on speech emotion recognition can help the model better understand contextual cues, speaker intentions, and subtle linguistic nuances, which are crucial for detecting sarcasm. Additionally, given that the SER is a more widely explored task with a comparatively larger availability of datasets, fine-tuning can help leverage knowledge and representations learned during pre-training providing improved generalization that can benefit sarcasm detection, especially when dealing with diverse and varied sarcastic speech patterns.

One of the challenges in fine-tuning pre-trained models is the discrepancy or mismatch between the domain in which the model was pre-trained and the target domain to which it is being applied. 
Task Adaptive Pre-Training [TAPT]~\cite{gururangan2020don}, is an existing NLP fine-tuning strategy that was proposed to resolve the domain shift by continuing the pre-training process on the target dataset.~\cite{chen2023exploring} introduced a novel fine-tuning method termed Psuedolabel-TAPT, which modifies the TAPT objective to learn contextualized emotion representations. Experiments show that P-TAPT performs better than TAPT, especially under low-resource settings. We adopt this strategy to fine-tune the existing wav2vec2 model on the IEMOCAP dataset for the SER task. The embeddings to be extracted for sarcasm detection are kept the same as the ones extracted above. We later provide an empirical analysis of both the pre- and post-fine-tuning features, as a part of the training for the sarcasm recognition task.

\subsection{Text Backbone: BART}
Bidirectional and Autoregressive Transformers (BART) ~\cite{lewis2019bart} is a denoising autoencoder designed to reconstruct the original document from a corrupted version. It employs a sequence-to-sequence model with a bidirectional encoder to process the corrupted text and a left-to-right autoregressive decoder. During pre-training, BART optimizes the negative log-likelihood of the uncorrupted document. Unlike other denoising autoencoders that are specifically designed for particular noise patterns, BART can handle various types of document corruption. In the most extreme scenario, where all source information is lost, BART functions as a language model. Additionally, as a sequence-to-sequence model, BART can be utilized for tasks such as text summarization, question answering, and translation.
BART generates a feature vector $x_t \in R^{d_t}$ for each instance x. To transform the text, we utilize the BART Large model with $d_t$ set to 1024. By taking the average of the representations from the last four transformer layers, we obtain a distinctive embedding representation for both the utterance and the context.

\section{Benchmarking \& Data Distribution}
\label{subsec:benchmarking}
We perform multi-class classification experiments by utilizing the features extracted as described above. Since we have three modalities, with- and without contextual information, we perform several ablation studies to understand the impact of the presence or absence of each of these aspects. To maintain a fair comparison with the prior literature, we adopt the same multimodal fusion mechanism used in~\cite{ray2022Mustardpp}, \ie a collaborative gated attention architecture~\cite{liu2019use}. Further, we perform experiments only for the speaker-independent setting, as we evaluate the performance of our models in a general setting, devoid of speaker names in the data. 

\paragraph{Training Details} We perform our experiments on a single NVIDIA RTX A5500 GPU where a single modality ablation takes $\sim 0.6$ hours, resulting in a total computation time of $20$ hours for the complete benchmarking reported in all tables. For each modality configuration, we report the average of $5$-fold cross-validation, where each fold is created using stratified K-fold separation, ensuring minimal skewness across the validation set in terms of the type of sarcasm. Each experiment is run for a maximum of $50$ epochs while deploying an early stopping mechanism with a patience of $5$ consecutive epochs unless the validation F1 score starts showing a non-increasing trend. We use $42$, $5141$, $1516$, $12667$ and $2238$ as random seeds for each training fold. We use a learning rate of $1e-3$, and a batch size of $256$ with the Adam optimizer with linear weight decay of $1e-2$.

\paragraph{Results} We perform experiments separately on the MUStARD++ and MUStARD++ with our extension separately, and report the average of 5 folds in Table~\ref{tab:bench1} and Table~\ref{tab:bench2} respectively. We contrast all the modality configurations in terms of mean Precision (P), mean Recall (R), and mean macro-F1, for both settings - with and without contextual information from the video clip. The labels in the first column depict the backbone architecture that was used to extract the features, more specifically, \textbf{T} - BART~\cite{lewis2019bart}, \textbf{T1} - Text encoder from ViFiCLIP, \textbf{V} - Average pooled intermediate ResNet\cite{ray2022Mustardpp}, \textbf{V1} - Video encoder from ViFiCLIP, \textbf{A} - Average pooled spectral features (MFCC) \cite{ray2022Mustardpp}, \textbf{A1} - wav2vec2, \textbf{A2} -wav2vec2 fine-tuned on SER. Values in \textbf{bold} show the best-performing modality combination, whereas the \underline{underlined} values highlight the cases where task performance was observed to be better without contextual information. For all other cases, \textit{models with contextual information outperform the models without contextual information}.
\begin{table}[!ht]
    \centering
    \begin{tabular}{cccc|ccc}
    \toprule
    \multirow{2}{*}{\textbf{Modal}} & \multicolumn{3}{c}{\textbf{Without Context}} & \multicolumn{3}{c}{\textbf{With Context}} \\
    & P & R & F1 & P & R & F1 \\
    \hline
    T & $0.679$ & $0.677$ & $0.677$ & $0.693$ & $0.692$ & $0.692$\\
    T1 & $0.692$ & $0.694$ & $0.692$ & $0.704$ & $0.713$ & $0.708$\\ \hline
    A & $0.639$ & $0.635$ & $0.636$ & $0.643$ & $0.641$ & $0.641$\\
    A1 & $0.652$ & $0.647$ & $0.649$ & $0.661$	& $0.668$ & $0.664$\\
    A2 & $0.664$ & $0.652$ & $0.657$ & $0.674$	& $0.678$ & $0.675$\\ \hline
    
    V & $0.595$	& $0.594$ & $0.594$ & $0.603$	& $0.600$ & $0.600$\\
    V1 & $0.615$ & $0.602$ & $0.608$ & $0.614$	& $0.612$ & $0.612$\\ \hline \hline
    
    T+A & $0.688$ & $0.686$ & $0.687$ & $0.702$	& $0.702$ & $0.702$\\
    T+A1 & $0.695$ & $0.698$ & $0.696$ & $0.705$ & $0.715$ & $0.709$\\
    T+A2 & $0.694$ & $0.705$ & $0.699$ & $0.712$ & $0.714$ & $0.712$\\
    T1+A1 & $0.715$ & $0.713$ & $0.713$ & $0.725$ & $0.725$ & $0.725$\\
    T1+A2 & $0.721$ & $0.725$ & $0.722$ & $0.729$ & $0.724$ & $0.726$\\ \hline

    A+V & $0.657$ & $0.654$ & $0.655$ & $0.675$ & $0.673$ & $0.674$\\
    A1+V & $0.671$ & $0.665$ & $0.667$ & $0.694$ & $0.682$ & $0.687$\\
    A2+V & $0.681$ & $0.668$ & $0.674$ & $0.705$ & $0.692$ & $0.698$\\
    A1+V1 & $0.698$ & $0.684$ & $0.690$ & $0.705$ & $0.709$ & $0.706$\\
    A2+V1 & $0.704$ & $0.691$ & $0.697$ & $0.714$ & $0.715$ & $0.714$\\ \hline
    
    V+T & $0.682$ & $0.681$ & \underline{$0.681$} & $0.679$ & $0.676$ & $0.676$\\
    V1+T & $0.692$ & $0.698$ & $0.694$ & $0.704$ & $0.691$ & $0.697$\\
    V1+T1 & $0.713$ & $0.712$ & $0.712$ & $0.718$ & $0.712$ & $0.714$\\ \hline \hline
    
    T+A+V & $0.695$ & $0.694$ & $0.694$ & $0.696$ & $0.695$ & $0.696$\\
    T+A1+V & $0.706$ & $0.709$ & $0.707$ & $0.712$ & $0.716$ & $0.713$\\
    T+A2+V & $0.705$ & $0.708$ & $0.706$ & $0.716$ & $0.718$ & $0.716$\\
    T+A1+V1 & $0.705$ & $0.718$ & $0.711$ & $0.718$ & $0.708$ & $0.712$\\
    T+A2+V1 & $0.718$ & $0.724$ & \underline{$0.720$} & $0.725$ & $0.714$ & $0.719$\\
    T1+A1+V & $0.724$ & $0.729$ & \underline{$0.726$} & $0.718$ & $0.724$ & $0.720$\\
    T1+A2+V & $0.724$ & $0.721$ & $0.722$ & $0.721$ & $0.732$ & $0.726$\\
    T1+A1+V1 & $0.734$ & $0.731$ & \underline{$0.732$} & $\mathbf{0.735}$ & $\mathbf{0.728}$ & $\mathbf{0.731}$\\
    T1+A2+V1 & $\mathbf{0.732}$ & $\mathbf{0.734}$ & \underline{$\mathbf{0.733}$} & $0.731$ & $0.729$ & $0.729$\\\hline
    \bottomrule
    \end{tabular}
    \vspace{0.5cm}
    \caption{On MUStARD++ Dataset. (Precision (P), mean Recall (R), and mean macro-F1 (primary evaluation metric))}
    \label{tab:bench1}
\end{table}

\begin{table}[!ht]
    \centering
    \begin{tabular}{cccc|ccc}
    \toprule
    \multirow{2}{*}{\textbf{Modal}} & \multicolumn{3}{c}{\textbf{Without Context}} & \multicolumn{3}{c}{\textbf{With Context}} \\
    & P & R & F1 & P & R & F1 \\
    \hline
    T1 & $0.714$ & $0.719$ & $0.716$ & $0.723$ & $0.715$ & $0.719$\\ \hline
    
    A1 & $0.682$	& $0.661$ & $0.671$ & $0.688$	& $0.691$ & $0.689$\\
    A2 & $0.675$	& $0.672$ & $0.673$ & $0.694$	& $0.695$ & $0.694$\\ \hline
    
    V1 & $0.617$	& $0.608$ & $0.612$ & $0.626$	& $0.619$ & $0.622$\\ \hline \hline

    T1+A1 & $0.701$ & $0.692$ & $0.696$ & $0.712$ & $0.712$ & $0.712$\\
    T1+A2 & $0.702$ & $0.706$ & $0.704$ & $0.726$ & $0.712$ & $0.719$\\\hline

    A1+V1 & $0.702$ & $0.691$ & $0.696$ & $0.713$ & $0.715$ & $0.714$\\
    A2+V1 & $0.715$ & $0.704$ & \underline{$0.709$} & $0.704$ & $0.709$ & $0.706$\\ \hline
    
    V1+T1 & $0.724$ & $0.711$ & $0.717$ & $0.726$ & $0.711$ & $0.718$\\ \hline\hline

    T1+A1+V1 & $\mathbf{0.728}$ & $\mathbf{0.737}$ & $\mathbf{0.732}$ & $0.737$ & $0.731$ & $0.734$\\
    T1+A2+V1 & $0.734$ & $0.726$ & $0.730$ & $\mathbf{0.738}$ & $\mathbf{0.735}$ & $\mathbf{0.736}$\\\hline

    \bottomrule
    \end{tabular}
    \vspace{0.5cm}
    \caption{On MUStARD++ Balanced Dataset}
    \label{tab:bench2}
\end{table}

\begin{table}[!ht]
    \centering
    \begin{tabular}{cccc}
         \toprule
         \textbf{Sarc Type} & \textbf{MUStARD++} & \textbf{Balanced} & $\%$ \textbf{Change} \\
         \midrule
         \textbf{PRO} & $0.705$ & $\mathbf{0.824}$ & $+11.77\%$  \\
         \textbf{ILL} & $0.484$ & $\mathbf{0.545}$ & $+6.06\%$  \\
         \textbf{EMB} & $0.696$ & $\mathbf{0.788}$ & $+9.09\%$  \\
         \textbf{NON} & $0.620$ & $\mathbf{0.720}$ & $+10.00\%$ \\\hline \hline
         \textbf{macro-F1} & $0.625$ & $\mathbf{0.720}$ & $\mathbf{+0.95}$ \\
         \bottomrule
    \end{tabular}
    \vspace{0.5cm}
    \caption{Distribution Shift helps task performance for sarcasm types and non-sarcastic (NON) instances.}
    \label{tab:bench3}
\end{table}

For the MUStARD++ dataset, it is clearly evident that the proposed benchmarking is superior and more robust compared to the prior work~\cite{ray2022Mustardpp}, \ie $T$, $A$ and $V$ backbones, where we surpass each unimodality with $1.59\%$, $2.19\%$ and $1.44\%$ across text, audio and video respectively. Moreover, upon fusing all modalities, we achieve a $3.89\%$ and $3.39\%$ improvement against the SOTA - $T+A+V$, when training utterances without context and 
with context, respectively. With the new encoders from pre-trained models, we report SOTA scores on MUStARD++.

When comparing the individual audio backbones, we find that the embeddings derived from the wav2vec2 model fine-tuned on the SER task ($A2$), surpass both the spectral features ($A$) and the ones derived from the wav2vec2 model trained on an ASR task ($A1$), by a substantial margin. Likewise, the video encoder ($V1$) and the text encoder ($T1$) from the pre-trained ViFiCLIP model outperform both their corresponding unimodal prior arts ($V$ and $T$, respectively). We hypothesize this is primarily because of the contrastive objective utilized in the training of ViFiCLIP and the large-scale training data that has been exposed to the model as a part of the CLIP pre-training.

Among all three modalities, the textual embeddings are more indicative of the sarcasm detection task than the audio and video embeddings. For audio, it is mainly owed to the mismatch between clean speech that was used to pre-train the wav2vec2 model (and even after fine-tuning on the SER task) and the noisy speech often superimposed with the background laughter, which is in a clear mismatch with the training data used for training the wav2vec2 model, or its fine-tuned model on the SER task. 

A similar trend is observed when evaluating our approach on the MUStARD++ dataset with our proposed extension, wherein the highest macro F1-scores of $0.732$ and $0.736$ are obtained by combining all three modalities, when training utterances without context and with context, respectively. 

\subsection{Exploring the Distribution shift}

To quantify the actual distribution shift introduced as a result of our extension, we compare the $V1+T1+A2$ configured model, trained with and without the extended annotations on a common test set. Formally, we segment out two sets, of 50 and 200 samples from the original MUStARD++ dataset, ensuring an equal distribution amongst all sarcasm types, and use it as our hold-out validation set and test set, respectively.

We showcase this in Table \ref{tab:bench3} where we first, train a model on the remainder of the $952$ samples from the MUStARD++ set, and report class-wise accuracies on the hold-out test set. Next, we replace randomly chosen samples from the training set of $952$ samples with samples from our proposed extension. We repeat this for $5$ random runs and report class-wise accuracies on the held-out test set while utilizing majority voting from each of the $5$ models for inference. Please note that the number of training samples is the same in both training setups.

It is clearly evident that merging the samples with our proposed extension helps improve the individual class-wise accuracies, and, eventually, an absolute improvement in the mean macro-F1 score of 0.95. Interestingly, apart from the improvement in the accuracies of three prominent types of sarcasm, namely, Propositional, Illocutionary and Embedded, we also find an improvement in the accuracy of non-sarcastic utterances. We hypothesize this is particularly because (a) intra-cluster variances among sarcastic types are now more defined, and (b) inter-cluster deviation between sarcastic and non-sarcastic is more well separated, owing to well-defined intra-clusters in the shared latent space of our model, and the addition of new non-sarcastic instances. 

\subsection*{Discussion}
\begin{figure*}[t!]
    \centering
    \begin{subfigure}[t]{0.29\textwidth}
        \centering
        \includegraphics[height=1.1in]{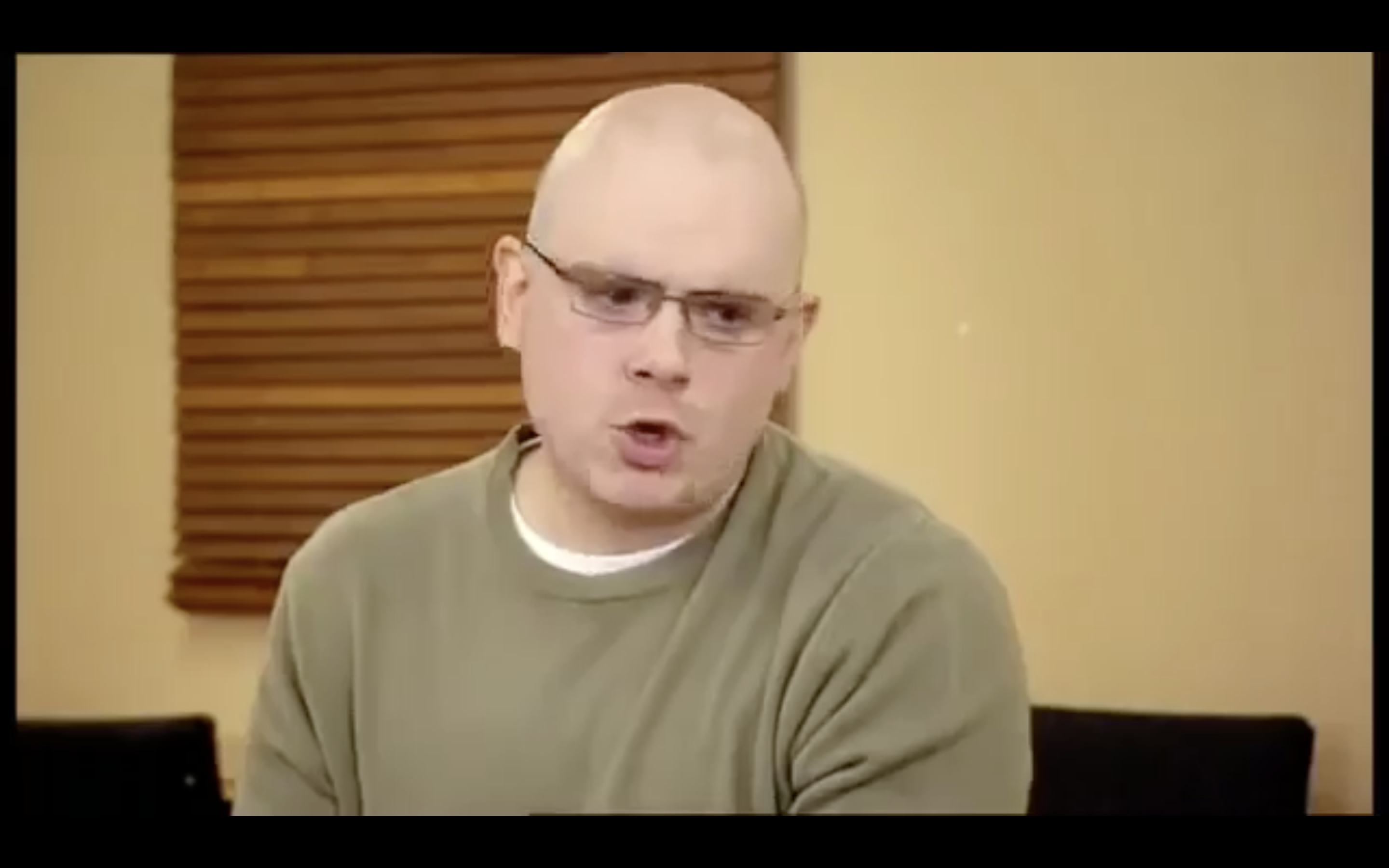}
        \caption{Nooo}
    \end{subfigure}%
    ~ 
    \begin{subfigure}[t]{0.29\textwidth}
        \centering
        \includegraphics[height=1.1in]{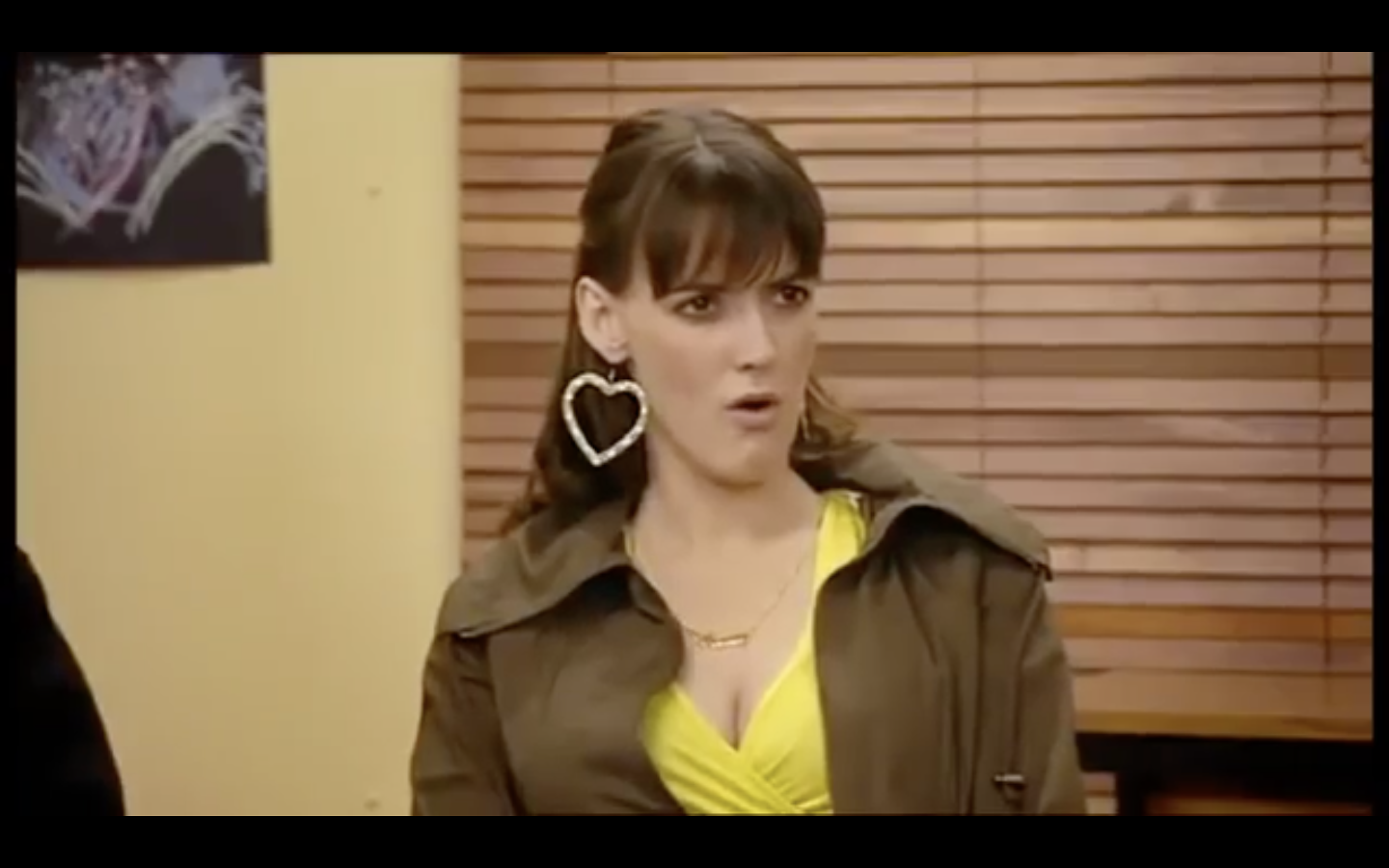}
        \caption{Nooo}
    \end{subfigure}
    ~ 
    \begin{subfigure}[t]{0.28\textwidth}
        \centering
        \includegraphics[height=1.1in]{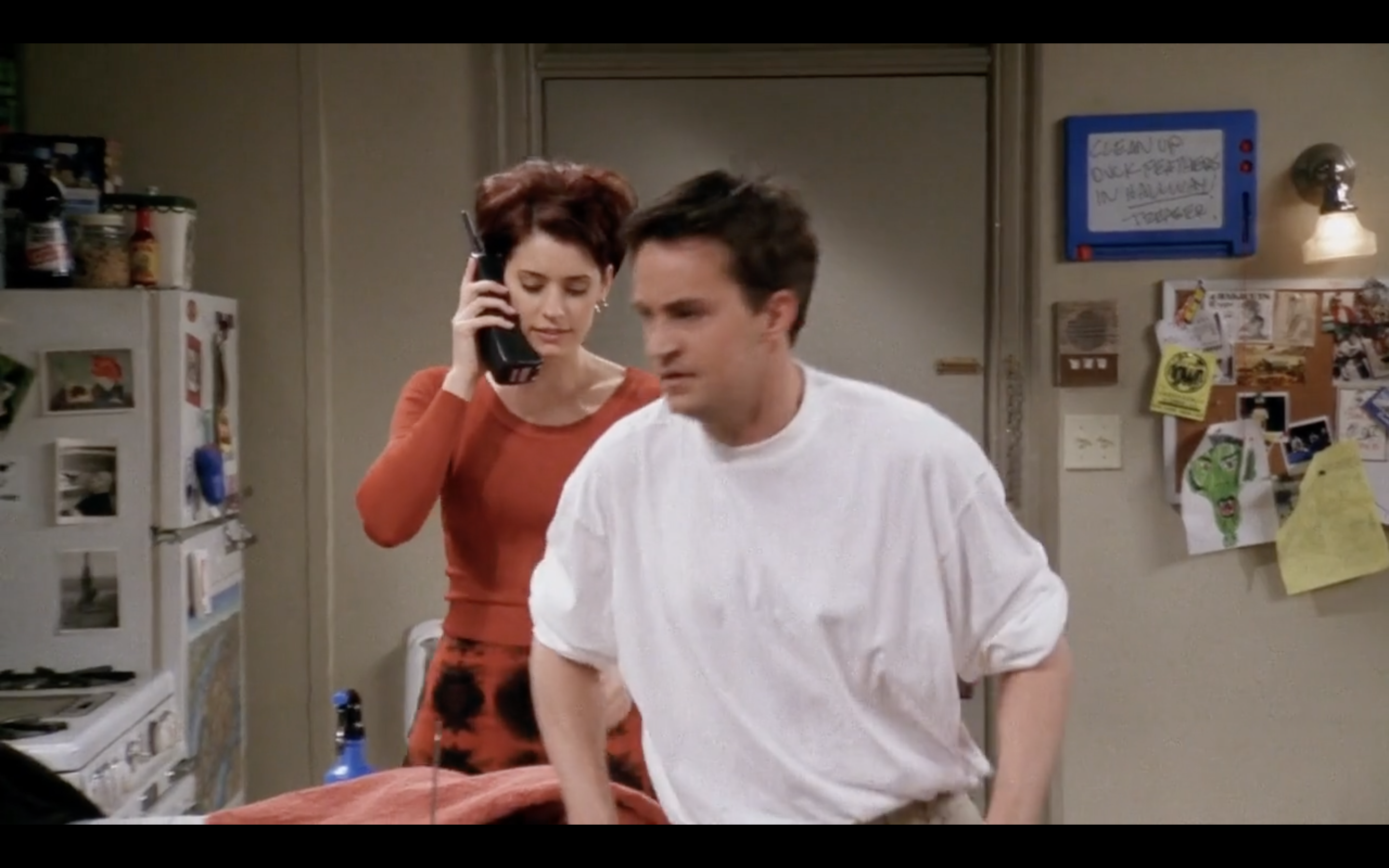}
        \caption{Hey ...}
    \end{subfigure}
    \caption{Mis-classified samples. Ground truth: Sarcastic; Type: Illocutionary.}
    \label{fig:fig1}
\end{figure*}

Among the different types of sarcasm discussed above, it has been argued in the literature that Illocutionary sarcasm is the most challenging since it involves non-speech like body language and gestures. Additionally, we observe that the \textbf{majority of the samples mis-classified by our model are the ones with short duration ($<2$ seconds)}, and hence do not contain enough context, to be processed by any of our backbones. In the captions for Figure~\ref{fig:fig1}, we show the utterances which are very limited in length and on the basis of which our models tried to predict sarcasm. 

Further, we do not perform the distribution shift experiment for Like-prefixed sarcasm we did not have enough samples for a test set. However, our experiment does reveal that non-sarcastic instances also benefit from instance addition. 

\section{Conclusion}

We perform comprehensive benchmarking on a multimodal dataset for the task of sarcasm detection. Our benchmarking reveals that the use of better encoders for both audio and video signals, improves the task performance by $3.39$, and $3.89$ percent points, with and without contextual information, respectively. We utilize pre-trained models by extracting features and training classification models in a supervised setting, with an existing feature fusion methodology. Additionally, we perform a quantitative analysis of the results obtained from various modality combinations, achieve SOTA performance on the dataset, and observe that Illocutionary sarcasm is more challenging to predict.

Our work also analyzes the model for sarcasm type imbalance and slightly alleviates it by adding new instances to the dataset. We call this proposed extension, \textit{MUStARD++ Balanced}. We validate this dataset extenuation with a substantial agreement based on Fleiss' Kappa and Krippendorf's alpha. Our recorded annotation experience and observations will help future extensions for this dataset, with a focus on finding instances of Like-prefixed sarcasm. Further, we choose the best modality combinations from unimodal, bimodal, and trimodal settings, and perform similar experiments to observe further improvement in the task performance. Our experiments on investigating the distribution shift for sarcasm types reveal that each sarcasm type observes an improvement in task performance by significant margins. Further, we perform a qualitative analysis of erroneous results from our experiments with the best-performing approach. We discuss our analysis and try to interpret why they fail to predict sarcasm correctly. 

In future, we plan to keep extending the dataset with new instances, especially for Illocutionary and Like-prefixed sarcasm types, and experiment with more feature fusion approaches. In the short term, we plan to identify instances which are $<2$ seconds in length and prune them from the data as they do not have sufficient context. We would like to further investigate linguistic phenomena like sarcasm, humour, and emotion in a more coherent setting, \ie multi-task learning-based setup where a joint model trains on three or more tasks under this \textit{sentiment umbrella}.

\section*{Limitations}
\label{sec:limit}

We acknowledge that our dataset extension is obtained from a publicly available source but we do release this data in the form of trimmed clips under a non-commercial license akin to existing datasets already existing in the same domain. Further, we would like to acknowledge that pre-trained models contain bias against certain members of society and can embed these biases in our models. We have already discussed the potential negative impacts of our models above and acknowledge that despite significantly improved task performance, they may not be able to identify such biases and wrongful predictions. We release the data, the code and the models only for the purpose of furthering research in this domain. 

\bibliographystyle{IEEEtran}
\bibliography{mybib}

\begin{thebibliography}{10}
\providecommand{\url}[1]{#1}
\csname url@samestyle\endcsname
\providecommand{\newblock}{\relax}
\providecommand{\bibinfo}[2]{#2}
\providecommand{\BIBentrySTDinterwordspacing}{\spaceskip=0pt\relax}
\providecommand{\BIBentryALTinterwordstretchfactor}{4}
\providecommand{\BIBentryALTinterwordspacing}{\spaceskip=\fontdimen2\font plus
\BIBentryALTinterwordstretchfactor\fontdimen3\font minus \fontdimen4\font\relax}
\providecommand{\BIBforeignlanguage}[2]{{%
\expandafter\ifx\csname l@#1\endcsname\relax
\typeout{** WARNING: IEEEtran.bst: No hyphenation pattern has been}%
\typeout{** loaded for the language `#1'. Using the pattern for}%
\typeout{** the default language instead.}%
\else
\language=\csname l@#1\endcsname
\fi
#2}}
\providecommand{\BIBdecl}{\relax}
\BIBdecl

\bibitem{Castro2019Mustard}
S.~Castro, D.~Hazarika, V.~P{\'e}rez-Rosas, R.~Zimmermann, R.~Mihalcea, and S.~Poria, ``Towards multimodal sarcasm detection (an {\_}{O}bviously{\_} perfect paper),'' in \emph{Proceedings of the 57th Annual Meeting of the Association for Computational Linguistics}, 2019.

\bibitem{ray2022Mustardpp}
A.~Ray, S.~Mishra, A.~Nunna, and P.~Bhattacharyya, ``A multimodal corpus for emotion recognition in sarcasm,'' in \emph{Proceedings of the Thirteenth Language Resources and Evaluation Conference}.\hskip 1em plus 0.5em minus 0.4em\relax Marseille, France: European Language Resources Association, 2022.

\bibitem{soleymani2017survey}
M.~Soleymani, D.~Garcia, B.~Jou, B.~Schuller, S.-F. Chang, and M.~Pantic, ``A survey of multimodal sentiment analysis,'' \emph{Image and Vision Computing}, vol.~65, pp. 3--14, 2017.

\bibitem{kiela2020hateful}
D.~Kiela, H.~Firooz, A.~Mohan, V.~Goswami, A.~Singh, P.~Ringshia, and D.~Testuggine, ``The hateful memes challenge: Detecting hate speech in multimodal memes,'' \emph{Advances in Neural Information Processing Systems}, vol.~33, pp. 2611--2624, 2020.

\bibitem{suryawanshi2020multimodal}
S.~Suryawanshi, B.~R. Chakravarthi, M.~Arcan, and P.~Buitelaar, ``Multimodal meme dataset (multioff) for identifying offensive content in image and text,'' in \emph{Proceedings of the second workshop on trolling, aggression and cyberbullying}, 2020, pp. 32--41.

\bibitem{guo2023audio}
P.~Guo, Z.~Chen, Y.~Li, and H.~Liu, ``Audio-visual fusion network based on conformer for multimodal emotion recognition,'' in \emph{Artificial Intelligence: Second CAAI International Conference, CICAI 2022, Beijing, China, August 27--28, 2022, Revised Selected Papers, Part II}.\hskip 1em plus 0.5em minus 0.4em\relax Springer, 2023, pp. 315--326.

\bibitem{liu2019use}
Y.~Liu, S.~Albanie, A.~Nagrani, and A.~Zisserman, ``Use what you have: Video retrieval using representations from collaborative experts,'' \emph{arXiv preprint arXiv:1907.13487}, 2019.

\bibitem{McFee2015Librosa}
B.~McFee, C.~Raffel, D.~Liang, D.~P. Ellis, M.~McVicar, E.~Battenberg, and O.~Nieto, ``librosa: Audio and music signal analysis in python.'' in \emph{Proceedings of the 14th python in science conference}, 2015.

\bibitem{opensmile}
``Opensmile,'' \url{https://audeering.github.io/opensmile/}.

\bibitem{He2016ResNet}
K.~He, X.~Zhang, S.~Ren, and J.~Sun, ``Deep residual learning for image recognition,'' in \emph{CVPR}, June 2016.

\bibitem{radford2021clip}
A.~Radford, J.~W. Kim, C.~Hallacy, A.~Ramesh, G.~Goh, S.~Agarwal, G.~Sastry, A.~Askell, P.~Mishkin, J.~Clark, G.~Krueger, and I.~Sutskever, ``Learning transferable visual models from natural language supervision,'' in \emph{ICML}, 2021.

\bibitem{rasheed2023fine}
H.~Rasheed, M.~U. Khattak, M.~Maaz, S.~Khan, and F.~S. Khan, ``Fine-tuned clip models are efficient video learners,'' in \emph{Proceedings of the IEEE/CVF Conference on Computer Vision and Pattern Recognition}, 2023, pp. 6545--6554.

\bibitem{NIPS2017_3f5ee243}
\BIBentryALTinterwordspacing
A.~Vaswani, N.~Shazeer, N.~Parmar, J.~Uszkoreit, L.~Jones, A.~N. Gomez, L.~u. Kaiser, and I.~Polosukhin, ``Attention is all you need,'' in \emph{Advances in Neural Information Processing Systems}, I.~Guyon, U.~V. Luxburg, S.~Bengio, H.~Wallach, R.~Fergus, S.~Vishwanathan, and R.~Garnett, Eds., vol.~30.\hskip 1em plus 0.5em minus 0.4em\relax Curran Associates, Inc., 2017. [Online]. Available: \url{https://proceedings.neurips.cc/paper_files/paper/2017/file/3f5ee243547dee91fbd053c1c4a845aa-Paper.pdf}
\BIBentrySTDinterwordspacing

\bibitem{10.1145/3124420}
\BIBentryALTinterwordspacing
A.~Joshi, P.~Bhattacharyya, and M.~J. Carman, ``Automatic sarcasm detection: A survey,'' \emph{ACM Comput. Surv.}, vol.~50, no.~5, sep 2017. [Online]. Available: \url{https://doi.org/10.1145/3124420}
\BIBentrySTDinterwordspacing

\bibitem{7549041}
M.~Bouazizi and T.~Otsuki~Ohtsuki, ``A pattern-based approach for sarcasm detection on twitter,'' \emph{IEEE Access}, vol.~4, pp. 5477--5488, 2016.

\bibitem{reyes-rosso-2011-mining}
\BIBentryALTinterwordspacing
A.~Reyes and P.~Rosso, ``Mining subjective knowledge from customer reviews: A specific case of irony detection,'' in \emph{Proceedings of the 2nd Workshop on Computational Approaches to Subjectivity and Sentiment Analysis ({WASSA} 2.011)}.\hskip 1em plus 0.5em minus 0.4em\relax Portland, Oregon: Association for Computational Linguistics, Jun. 2011, pp. 118--124. [Online]. Available: \url{https://aclanthology.org/W11-1715}
\BIBentrySTDinterwordspacing

\bibitem{barbieri-etal-2014-modelling}
\BIBentryALTinterwordspacing
F.~Barbieri, H.~Saggion, and F.~Ronzano, ``Modelling sarcasm in {T}witter, a novel approach,'' in \emph{Proceedings of the 5th Workshop on Computational Approaches to Subjectivity, Sentiment and Social Media Analysis}.\hskip 1em plus 0.5em minus 0.4em\relax Baltimore, Maryland: Association for Computational Linguistics, Jun. 2014, pp. 50--58. [Online]. Available: \url{https://aclanthology.org/W14-2609}
\BIBentrySTDinterwordspacing

\bibitem{van-hee-etal-2018-semeval}
\BIBentryALTinterwordspacing
C.~Van~Hee, E.~Lefever, and V.~Hoste, ``{S}em{E}val-2018 task 3: Irony detection in {E}nglish tweets,'' in \emph{Proceedings of the 12th International Workshop on Semantic Evaluation}.\hskip 1em plus 0.5em minus 0.4em\relax New Orleans, Louisiana: Association for Computational Linguistics, Jun. 2018, pp. 39--50. [Online]. Available: \url{https://aclanthology.org/S18-1005}
\BIBentrySTDinterwordspacing

\bibitem{Potamias_2020}
\BIBentryALTinterwordspacing
R.~A. Potamias, G.~Siolas, and A.~G. Stafylopatis, ``A transformer-based approach to irony and sarcasm detection,'' \emph{Neural Computing and Applications}, vol.~32, no.~23, pp. 17\,309--17\,320, jun 2020. [Online]. Available: \url{https://doi.org/10.1007%2Fs00521-020-05102-3}
\BIBentrySTDinterwordspacing

\bibitem{riloff-etal-2013-sarcasm}
\BIBentryALTinterwordspacing
E.~Riloff, A.~Qadir, P.~Surve, L.~De~Silva, N.~Gilbert, and R.~Huang, ``Sarcasm as contrast between a positive sentiment and negative situation,'' in \emph{Proceedings of the 2013 Conference on Empirical Methods in Natural Language Processing}.\hskip 1em plus 0.5em minus 0.4em\relax Seattle, Washington, USA: Association for Computational Linguistics, Oct. 2013, pp. 704--714. [Online]. Available: \url{https://aclanthology.org/D13-1066}
\BIBentrySTDinterwordspacing

\bibitem{shangipour-ataei-etal-2020-applying}
\BIBentryALTinterwordspacing
T.~Shangipour~ataei, S.~Javdan, and B.~Minaei-Bidgoli, ``Applying transformers and aspect-based sentiment analysis approaches on sarcasm detection,'' in \emph{Proceedings of the Second Workshop on Figurative Language Processing}.\hskip 1em plus 0.5em minus 0.4em\relax Online: Association for Computational Linguistics, Jul. 2020, pp. 67--71. [Online]. Available: \url{https://aclanthology.org/2020.figlang-1.9}
\BIBentrySTDinterwordspacing

\bibitem{devlin2019bert}
J.~Devlin, M.-W. Chang, K.~Lee, and K.~Toutanova, ``Bert: Pre-training of deep bidirectional transformers for language understanding,'' 2019.

\bibitem{ijcai2017p568}
\BIBentryALTinterwordspacing
D.~Ma, S.~Li, X.~Zhang, and H.~Wang, ``Interactive attention networks for aspect-level sentiment classification,'' in \emph{Proceedings of the Twenty-Sixth International Joint Conference on Artificial Intelligence, {IJCAI-17}}, 2017, pp. 4068--4074. [Online]. Available: \url{https://doi.org/10.24963/ijcai.2017/568}
\BIBentrySTDinterwordspacing

\bibitem{abu-farha-magdy-2021-benchmarking}
\BIBentryALTinterwordspacing
I.~Abu~Farha and W.~Magdy, ``Benchmarking transformer-based language models for {A}rabic sentiment and sarcasm detection,'' in \emph{Proceedings of the Sixth Arabic Natural Language Processing Workshop}.\hskip 1em plus 0.5em minus 0.4em\relax Kyiv, Ukraine (Virtual): Association for Computational Linguistics, Apr. 2021, pp. 21--31. [Online]. Available: \url{https://aclanthology.org/2021.wanlp-1.3}
\BIBentrySTDinterwordspacing

\bibitem{conneau-etal-2020-unsupervised}
\BIBentryALTinterwordspacing
A.~Conneau, K.~Khandelwal, N.~Goyal, V.~Chaudhary, G.~Wenzek, F.~Guzm{\'a}n, E.~Grave, M.~Ott, L.~Zettlemoyer, and V.~Stoyanov, ``Unsupervised cross-lingual representation learning at scale,'' in \emph{Proceedings of the 58th Annual Meeting of the Association for Computational Linguistics}.\hskip 1em plus 0.5em minus 0.4em\relax Online: Association for Computational Linguistics, Jul. 2020, pp. 8440--8451. [Online]. Available: \url{https://aclanthology.org/2020.acl-main.747}
\BIBentrySTDinterwordspacing

\bibitem{abdulmageed2021arbert}
M.~Abdul-Mageed, A.~Elmadany, and E.~M.~B. Nagoudi, ``Arbert \& marbert: Deep bidirectional transformers for arabic,'' 2021.

\bibitem{pennington-etal-2014-glove}
\BIBentryALTinterwordspacing
J.~Pennington, R.~Socher, and C.~Manning, ``{G}lo{V}e: Global vectors for word representation,'' in \emph{Proceedings of the 2014 Conference on Empirical Methods in Natural Language Processing ({EMNLP})}.\hskip 1em plus 0.5em minus 0.4em\relax Doha, Qatar: Association for Computational Linguistics, Oct. 2014, pp. 1532--1543. [Online]. Available: \url{https://aclanthology.org/D14-1162}
\BIBentrySTDinterwordspacing

\bibitem{goel2022sarcasm}
P.~Goel, R.~Jain, A.~Nayyar, S.~Singhal, and M.~Srivastava, ``Sarcasm detection using deep learning and ensemble learning,'' \emph{Multimedia Tools and Applications}, vol.~81, no.~30, pp. 43\,229--43\,252, 2022.

\bibitem{khodak2018large}
M.~Khodak, N.~Saunshi, and K.~Vodrahalli, ``A large self-annotated corpus for sarcasm,'' 2018.

\bibitem{perez2013utterance}
V.~P{\'e}rez-Rosas, R.~Mihalcea, and L.-P. Morency, ``Utterance-level multimodal sentiment analysis,'' in \emph{Proceedings of the 51st Annual Meeting of the Association for Computational Linguistics (Volume 1: Long Papers)}, 2013, pp. 973--982.

\bibitem{zadeh2016mosi}
A.~Zadeh, R.~Zellers, E.~Pincus, and L.-P. Morency, ``Mosi: multimodal corpus of sentiment intensity and subjectivity analysis in online opinion videos,'' \emph{arXiv preprint arXiv:1606.06259}, 2016.

\bibitem{busso2008iemocap}
C.~Busso, M.~Bulut, C.-C. Lee, A.~Kazemzadeh, E.~Mower, S.~Kim, J.~N. Chang, S.~Lee, and S.~S. Narayanan, ``Iemocap: Interactive emotional dyadic motion capture database,'' \emph{Language resources and evaluation}, vol.~42, no.~4, pp. 335--359, 2008.

\bibitem{poria2017context}
S.~Poria, E.~Cambria, D.~Hazarika, N.~Majumder, A.~Zadeh, and L.-P. Morency, ``Context-dependent sentiment analysis in user-generated videos,'' in \emph{Proceedings of the 55th annual meeting of the association for computational linguistics (volume 1: Long papers)}, 2017, pp. 873--883.

\bibitem{majumder2018multimodal}
N.~Majumder, D.~Hazarika, A.~Gelbukh, E.~Cambria, and S.~Poria, ``Multimodal sentiment analysis using hierarchical fusion with context modeling,'' \emph{Knowledge-based systems}, vol. 161, pp. 124--133, 2018.

\bibitem{hazarika-etal-2018-conversational}
\BIBentryALTinterwordspacing
D.~Hazarika, S.~Poria, A.~Zadeh, E.~Cambria, L.-P. Morency, and R.~Zimmermann, ``Conversational memory network for emotion recognition in dyadic dialogue videos,'' in \emph{Proceedings of the 2018 Conference of the North {A}merican Chapter of the Association for Computational Linguistics: Human Language Technologies, Volume 1 (Long Papers)}.\hskip 1em plus 0.5em minus 0.4em\relax New Orleans, Louisiana: Association for Computational Linguistics, Jun. 2018, pp. 2122--2132. [Online]. Available: \url{https://aclanthology.org/N18-1193}
\BIBentrySTDinterwordspacing

\bibitem{wang2019words}
Y.~Wang, Y.~Shen, Z.~Liu, P.~P. Liang, A.~Zadeh, and L.-P. Morency, ``Words can shift: Dynamically adjusting word representations using nonverbal behaviors,'' in \emph{Proceedings of the AAAI Conference on Artificial Intelligence}, vol.~33, no.~01, 2019, pp. 7216--7223.

\bibitem{pham2019found}
H.~Pham, P.~P. Liang, T.~Manzini, L.-P. Morency, and B.~P{\'o}czos, ``Found in translation: Learning robust joint representations by cyclic translations between modalities,'' in \emph{Proceedings of the AAAI Conference on Artificial Intelligence}, vol.~33, no.~01, 2019, pp. 6892--6899.

\bibitem{liang2018multimodal}
P.~P. Liang, Z.~Liu, A.~B. Zadeh, and L.-P. Morency, ``Multimodal language analysis with recurrent multistage fusion,'' in \emph{Proceedings of the 2018 Conference on Empirical Methods in Natural Language Processing}, 2018, pp. 150--161.

\bibitem{tsai2018learning}
Y.-H.~H. Tsai, P.~P. Liang, A.~Zadeh, L.-P. Morency, and R.~Salakhutdinov, ``Learning factorized multimodal representations,'' \emph{arXiv preprint arXiv:1806.06176}, 2018.

\bibitem{sahay-etal-2020-low}
\BIBentryALTinterwordspacing
S.~Sahay, E.~Okur, S.~H~Kumar, and L.~Nachman, ``Low rank fusion based transformers for multimodal sequences,'' in \emph{Second Grand-Challenge and Workshop on Multimodal Language (Challenge-HML)}.\hskip 1em plus 0.5em minus 0.4em\relax Seattle, USA: Association for Computational Linguistics, Jul. 2020, pp. 29--34. [Online]. Available: \url{https://aclanthology.org/2020.challengehml-1.4}
\BIBentrySTDinterwordspacing

\bibitem{tsai-etal-2019-multimodal}
\BIBentryALTinterwordspacing
Y.-H.~H. Tsai, S.~Bai, P.~P. Liang, J.~Z. Kolter, L.-P. Morency, and R.~Salakhutdinov, ``Multimodal transformer for unaligned multimodal language sequences,'' in \emph{Proceedings of the 57th Annual Meeting of the Association for Computational Linguistics}.\hskip 1em plus 0.5em minus 0.4em\relax Florence, Italy: Association for Computational Linguistics, Jul. 2019, pp. 6558--6569. [Online]. Available: \url{https://aclanthology.org/P19-1656}
\BIBentrySTDinterwordspacing

\bibitem{liu-etal-2018-efficient-low}
\BIBentryALTinterwordspacing
Z.~Liu, Y.~Shen, V.~B. Lakshminarasimhan, P.~P. Liang, A.~Bagher~Zadeh, and L.-P. Morency, ``Efficient low-rank multimodal fusion with modality-specific factors,'' in \emph{Proceedings of the 56th Annual Meeting of the Association for Computational Linguistics (Volume 1: Long Papers)}.\hskip 1em plus 0.5em minus 0.4em\relax Melbourne, Australia: Association for Computational Linguistics, Jul. 2018, pp. 2247--2256. [Online]. Available: \url{https://aclanthology.org/P18-1209}
\BIBentrySTDinterwordspacing

\bibitem{9753058}
A.~Bhat and A.~Chauhan, ``Multimodal sarcasm detection: A survey,'' in \emph{2022 IEEE Delhi Section Conference (DELCON)}, 2022, pp. 1--7.

\bibitem{mustard}
S.~Castro, D.~Hazarika, V.~P{\'e}rez-Rosas, R.~Zimmermann, R.~Mihalcea, and S.~Poria, ``Towards multimodal sarcasm detection (an \_obviously\_ perfect paper),'' in \emph{Proceedings of the 57th Annual Meeting of the Association for Computational Linguistics (Volume 1: Long Papers)}.\hskip 1em plus 0.5em minus 0.4em\relax Florence, Italy: Association for Computational Linguistics, 7 2019.

\bibitem{ray-etal-2022-multimodal}
\BIBentryALTinterwordspacing
A.~Ray, S.~Mishra, A.~Nunna, and P.~Bhattacharyya, ``A multimodal corpus for emotion recognition in sarcasm,'' in \emph{Proceedings of the Thirteenth Language Resources and Evaluation Conference}.\hskip 1em plus 0.5em minus 0.4em\relax Marseille, France: European Language Resources Association, Jun. 2022, pp. 6992--7003. [Online]. Available: \url{https://aclanthology.org/2022.lrec-1.756}
\BIBentrySTDinterwordspacing

\bibitem{liang2021multibench}
P.~P. Liang, Y.~Lyu, X.~Fan, Z.~Wu, Y.~Cheng, J.~Wu, L.~Y. Chen, P.~Wu, M.~A. Lee, Y.~Zhu \emph{et~al.}, ``Multibench: Multiscale benchmarks for multimodal representation learning,'' in \emph{Thirty-fifth Conference on Neural Information Processing Systems Datasets and Benchmarks Track (Round 1)}, 2021.

\bibitem{chen2023exploring}
L.-W. Chen and A.~Rudnicky, ``Exploring wav2vec 2.0 fine tuning for improved speech emotion recognition,'' in \emph{ICASSP 2023-2023 IEEE International Conference on Acoustics, Speech and Signal Processing (ICASSP)}.\hskip 1em plus 0.5em minus 0.4em\relax IEEE, 2023, pp. 1--5.

\bibitem{Chauhan2020SarcasmEmotion}
D.~S. Chauhan, D.~S~R, A.~Ekbal, and P.~Bhattacharyya, ``Sentiment and emotion help sarcasm? a multi-task learning framework for multi-modal sarcasm, sentiment and emotion analysis,'' in \emph{Proceedings of the 58th Annual Meeting of the Association for Computational Linguistics}, 2020.

\bibitem{picard1997affective}
R.~W. Picard, ``Affective computing mit press,'' \emph{Cambridge, Massachsusetts}, p.~2, 1997.

\bibitem{gururangan2020don}
S.~Gururangan, A.~Marasovi{\'c}, S.~Swayamdipta, K.~Lo, I.~Beltagy, D.~Downey, and N.~A. Smith, ``Don't stop pretraining: Adapt language models to domains and tasks,'' \emph{arXiv preprint arXiv:2004.10964}, 2020.

\bibitem{lewis2019bart}
M.~Lewis, Y.~Liu, N.~Goyal, M.~Ghazvininejad, A.~Mohamed, O.~Levy, V.~Stoyanov, and L.~Zettlemoyer, ``Bart: Denoising sequence-to-sequence pre-training for natural language generation, translation, and comprehension,'' \emph{arXiv preprint arXiv:1910.13461}, 2019.

\end{thebibliography}

\end{document}